\documentclass{article} 

\usepackage{natbib}
\usepackage{enumitem,amssymb}
\usepackage{amsmath}
\usepackage{amsfonts}
\usepackage{marginnote}
\usepackage{graphicx}
\usepackage[T1]{fontenc}
\usepackage{ifthen}
\usepackage{url}
\usepackage{fancyhdr}
\usepackage{hyperref}
\usepackage{lipsum}
\usepackage{xfrac}
\usepackage{booktabs}
\usepackage{tabularx}
\usepackage{pifont}
\usepackage{xcolor}
\usepackage{glossaries}
\glsdisablehyper
\usepackage{multirow}
\usepackage{xspace}
\usepackage{tikz}
\usepackage{xurl}
\usepackage[accepted]{icml2022}

\newcommand{\cmark}{\tikz[baseline=-0.5ex]\draw[black,fill=black,radius=2pt] (0,0) circle ;}

\newacronym{bpda}{BPDA}{Backward Pass Differentiable Approximation}
\newacronym{eot}{EoT}{Expectation over Transformation}
\newacronym{pgd}{PGD}{Projected Gradient Descent}
\newacronym{awp}{AWP}{Adversarial Weight Perturbation}
\newacronym{ebm}{EBM}{Energy-Based Model}
\newacronym{dsm}{DSM}{Denoising Score-Matching}
\newacronym{adp}{ADP}{Adaptive Denoising Purification}
\newacronym{apgd}{APGD}{AutoPGD}
\newacronym{fgsm}{FGSM}{Fast Gradient Sign Method}
\newacronym{pca}{PCA}{Principal Component Analysis}
\newacronym{mcmc}{MCMC}{Markov chain Monte Carlo}
\newacronym{hd}{HD}{Hedge Defense}
\newacronym{dlr}{DLR}{Difference of Logits Ratio}
\newacronym{ode}{ODE}{Ordinary Differential Equation}
\newacronym{cw}{CW}{Carlini-Wagner}
\newacronym{atld}{ATLD}{Adversarial Training with Latent Distribution}
\newacronym{imf}{IMF}{Inference with Manifold Transformation}

\newcommand{\cifar}{\textsc{Cifar-10}\xspace}

\newcommand{\cifarh}{\textsc{Cifar-100}\xspace}

\newcommand{\svhn}{\textsc{Svhn}\xspace}
\newcommand{\autoattack}{\textsc{AutoAttack}\xspace}
\newcommand{\robustbench}{\texttt{RobustBench}\xspace}
\newcommand{\sodef}{SODEF\xspace}
\newcommand{\trades}{TRADES\xspace}
\newcommand{\aidpurifier}{AID-purifier\xspace}
\newcommand{\aidpurified}{AID-purified\xspace}

\usepackage{amsmath,amsfonts,bm}

\def\eqref#1{equation~\ref{#1}}

\def\1{\bm{1}}
\newcommand{\train}{\mathcal{D}}

\def\vdelta{{\bm{\delta}}}
\def\vxi{{\bm{\xi}}}

\def\vu{{\bm{u}}}

\def\vx{{\bm{x}}}

\DeclareMathAlphabet{\mathsfit}{\encodingdefault}{\sfdefault}{m}{sl}
\SetMathAlphabet{\mathsfit}{bold}{\encodingdefault}{\sfdefault}{bx}{n}

\newcommand{\E}{\mathbb{E}}

\newcommand{\R}{\mathbb{R}}

\newcommand{\softmax}{\mathrm{softmax}}

\DeclareMathOperator*{\argmax}{arg\,max}
\DeclareMathOperator*{\argmin}{arg\,min}

\DeclareMathOperator*{\maximize}{max}

\def\R{\mathbb{R}}

\newcommand{\norm}[1]{\left\|#1\right\|}

\newcolumntype{C}[1]{>{\centering\arraybackslash}p{#1}}
\newcolumntype{L}[1]{>{\raggedright\arraybackslash}p{#1}}
\newcolumntype{R}[1]{>{\raggedleft\arraybackslash}p{#1}}

\icmltitlerunning{Evaluating the Adversarial Robustness of Adaptive Test-time Defenses}
\begin{document}

\twocolumn[
\icmltitle{Evaluating the Adversarial Robustness of Adaptive Test-time Defenses}



\icmlsetsymbol{equal}{*}

\begin{icmlauthorlist}
\icmlauthor{Francesco Croce}{equal,tu}
\icmlauthor{Sven Gowal}{equal,dm}
\icmlauthor{Thomas Brunner}{equal,edr}
\icmlauthor{Evan Shelhamer}{equal,dm}
\icmlauthor{Matthias Hein}{tu}
\icmlauthor{Taylan Cemgil}{dm}
\end{icmlauthorlist}

\icmlaffiliation{edr}{Everyday Robots, Munich, Germany}
\icmlaffiliation{tu}{University of T\"ubingen, Germany}
\icmlaffiliation{dm}{DeepMind, London, United Kingdom}

\icmlcorrespondingauthor{Francesco Croce}{francesco.croce@uni-tuebingen.de}
\icmlcorrespondingauthor{Sven Gowal, Evan Shelhamer}{\{sgowal,shelhamer\}@deepmind.com}
\icmlcorrespondingauthor{Thomas Brunner}{tbrunner@everydayrobots.com}

\icmlkeywords{Adversarial Robustness, Adaptation, Test-Time Defenses}

\vskip 0.3in
]



\printAffiliationsAndNotice{\icmlEqualContribution} 

\begin{abstract}
Adaptive defenses, which optimize at test time, promise to improve adversarial robustness.
We categorize such \emph{adaptive test-time defenses},
explain their potential benefits and drawbacks,
and evaluate a representative variety of the latest adaptive defenses for image classification.
Unfortunately, none significantly improve upon static defenses when subjected to our careful case study evaluation.
Some even weaken the underlying static model while simultaneously increasing inference computation.
While these results are disappointing, we still believe that adaptive test-time defenses are a promising avenue of research and, as such, we provide recommendations for their thorough evaluation.
We extend the checklist of \citet{carlini_evaluating_2019} by providing concrete steps specific to adaptive defenses.
\end{abstract}

\section{Introduction}

Despite the successes of deep learning \citep{goodfellow_deep_2016}, it is well-known that deep networks are not intrinsically robust.
In particular, the addition of small but carefully chosen $\ell_p$-norm bounded deviations to the input, called adversarial perturbations, can cause a deep network to make incorrect predictions with high confidence \citep{carlini_adversarial_2017,carlini_towards_2017,goodfellow_explaining_2014,kurakin_adversarial_2016,szegedy_intriguing_2013,biggio2013evasion}.
Starting with \citet{biggio2013evasion} and \citet{szegedy_intriguing_2013}, there has been a lot of work on understanding and generating adversarial perturbations \citep{carlini_towards_2017,athalye_synthesizing_2017,uesato_adversarial_2018}, and on building models that are robust to such perturbations \citep{goodfellow_explaining_2014,papernot_distillation_2015,madry_towards_2017,kannan_adversarial_2018,zhang_theoretically_2019,rice_overfitting_2020,gowal_uncovering_2020,gowal2021improving}.

A model $f$ robust to $\ell_p$-norm bounded perturbations aims to minimize the adversarial risk \citep{madry_towards_2017}
\begin{equation}
\E_{(\vx,y) \sim \train} \left[ \maximize_{\vdelta \in \{ \vdelta : \norm{\vdelta}_p < \epsilon \}} L(f(\vx + \vdelta), y) \right]
\label{eq:adversarial_risk}
\end{equation}
where $\mathcal{D}$ is a data distribution over pairs of examples $\vx$ and corresponding labels $y$, $L(z, y)$ is a suitable loss function (such as the 0-1 loss for classification), and $\epsilon$ is a maximal norm which defines the set of allowed perturbations for a given example $\vx$.
As such, finding the worst-case perturbation (or optimal adversarial perturbation) $\vdelta^\star$ is key for both training and testing models.
Finding sub-optimal $\hat{\vdelta}$ with $L(f(\vx + \hat{\vdelta}), y) \leq L(f(\vx + \vdelta^\star), y)$ estimates lower bounds on the true adversarial risk (or in the context of a classification task, upper bounds on the true robust accuracy), which may give a false sense of security.
Hence, it is practically important to optimally solve the problem in \autoref{eq:adversarial_risk}.

The difficulty of solving this maximization problem close to optimality is highlighted by \citet{uesato_adversarial_2018,athalye_obfuscated_2018,tramer_adaptive_2020}, in which a number of published defenses are easily broken by adaptive adversarial attacks that are tuned to the defenses.
This reversal highlights the importance of understanding the limitations of different robust training and testing methods \citep{pintor_indicators_2021}.
Standardizing evaluation allows for the systematic tracking of real progress on adversarial robustness.
However, to guarantee that standardized evaluations are accurate, defenses must adhere to some practical restrictions.
\robustbench~\citep{croce2020robustbench}, for example, rules out \emph{(i)} models which have zero gradients with respect to the input  \citep{buckman_thermometer_2018,guo_countering_2018}, \emph{(ii)} randomized models \citep{yang2019menet,pang2020mixup}, and \emph{(iii)} models that optimize during inference \citep{samangouei2018defensegan,li2019generative,schott2019towards}.
These restrictions may unnecessarily limit the development of robust models and, as a consequence, several researchers have offered general recommendations on how to evaluate adversarial defenses~\citep{carlini_evaluating_2019}.
However, the evaluation of each new defense raises non-trivial choices.

In this work, we focus on \emph{adaptive test-time defenses} that apply iterative optimization during inference,
which promise to circumvent limitations of \emph{static} defenses.
In the context of adversarial robustness, the optimization is designed to ``purify'' inputs before feeding them to a static model or to adapt the model itself (we elaborate on categorizations of adaptive test-time defenses in \autoref{sec:definition}).
We restrict our analysis to image classification because it is the most common test-bed for studying robustness against $\ell_p$-norm bounded attacks.

We foresee adaptive defenses as an important step towards building robust defenses.
However, given their novelty, the evaluation of such defenses has not been standardized.
Indeed, test-time optimization can prevent the proper operation of standardized attacks established for static defenses, such as
\autoattack~\citep{croce_reliable_2020}.
This optimization also renders approximations such as the \gls{bpda}~\citep{athalye_obfuscated_2018} more difficult to apply.
As a result, recent works
make strong robustness claims that turn out to be void or significantly weaker than claimed.
In this paper:
\begin{itemize}[leftmargin=0.4cm,noitemsep,topsep=0pt]
\item[\ding{182}] We categorize adaptive test-time defenses and explain their potential benefits and drawbacks;
\item[\ding{183}] We evaluate 9 recent defenses (see Table \ref{tab:summary_casestudy}), and show that they significantly overestimate their robustness.
  Relative to static defenses, most provide little improvement, and some are even detrimental. Furthermore all require more inference computation;
\item[\ding{184}] Finally, we provide recommendations for evaluating such defenses and explain under which circumstances standard evaluations, like
\autoattack, are accurate.
\end{itemize}

\section{Adaptive Test-Time Defenses}\label{sec:definition}

We provide a taxonomy of adaptive test-time defenses.
We highlight their operating principles in \autoref{sec:principles}, identify their building blocks in \autoref{sec:buildingblocks}, and elaborate on potential pitfalls in \autoref{sec:pitfalls}.
We also clearly delineate the threat model and adversarial capabilities that we assume during our case study of various techniques in \autoref{sec:threat}.

\subsection{Principles}\label{sec:principles}

\begin{figure}[t]
\centering
\includegraphics[width=\columnwidth]{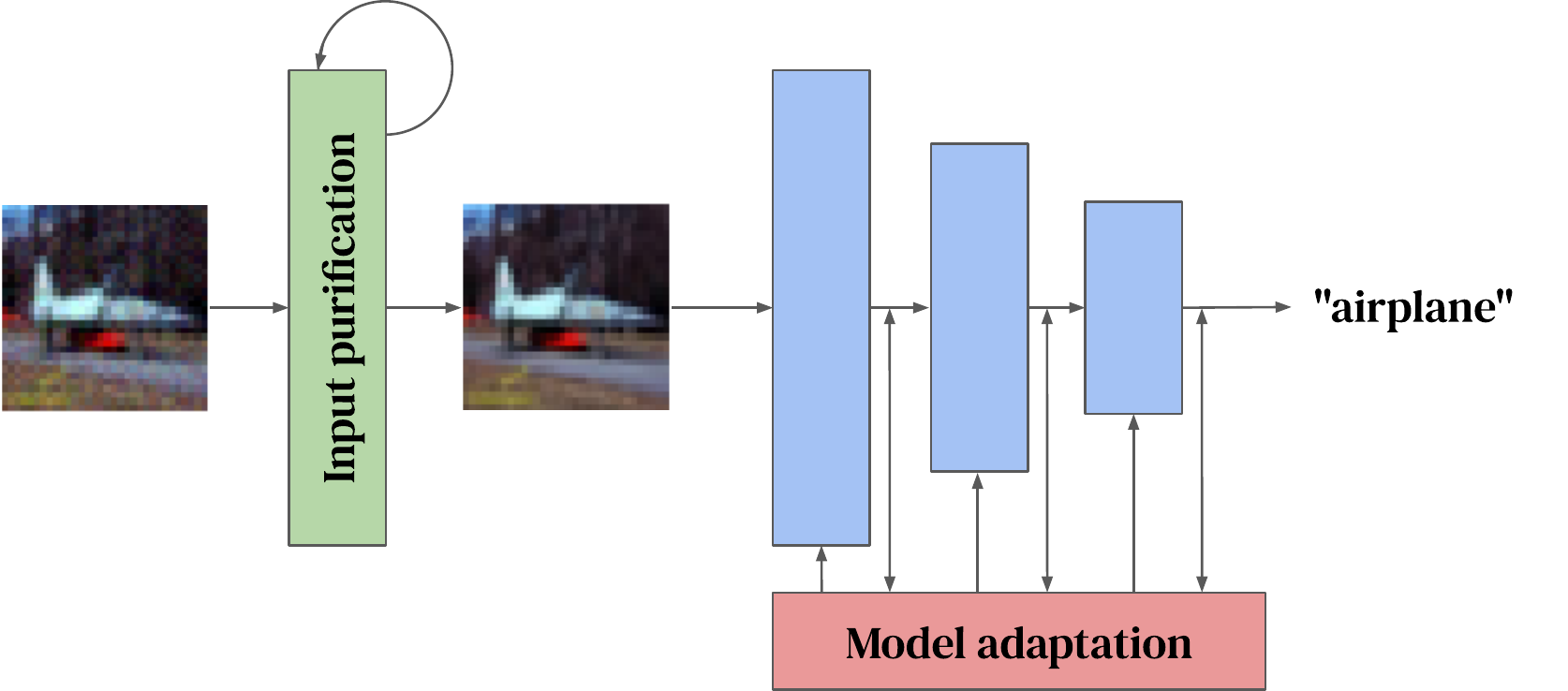}
\caption{
Adaptive test-time defenses operate by adapting their computation to the input.
A first category of adaptive test-time defenses operate in the input space and aim to ``purify'' inputs before they are fed to the standard pre-trained model.
A second category aims at adapting model parameters or intermediate activations.
}
\label{fig:architecture}
\end{figure}

Static defenses (e.g., standard deep networks) use a trained model for which the inputs and parameters are fixed at inference.
In contrast, adaptive test-time defenses can alter the input or parameters for test inputs, either by optimizing an auxiliary loss or conditioning directly on the input.
We identify two principles (highlighted in \autoref{fig:architecture}).
\begin{description}[leftmargin=0pt,itemsep=2pt,topsep=0pt,parsep=0pt]
\item[Input purification (IP):] A model (possibly pre-trained for robustness) is augmented with test-time optimization to alter its inputs before the model is applied.
The input optimization procedure may rely on hand-crafted \citep{alfarra2021combating,wu2021attacking} or learned objectives \citep{qian2021improving,mao2021adversarial,hwang2021aidpurifier} which may involve auxiliary networks such as generative models \citep{song2018pixeldefend,samangouei2018defensegan,yoon2021adversarial,nie2022diffusion}.
\item[Model adaptation (MA):] A model is made adaptive by applying an optimization procedure to alter its parameters or state (e.g., batch normalization statistics) \citep{wang2021fighting} or activations \citep{chen2021towards} during inference.
Other iterative schemes, such as implicit layers \citep{kang2021stable}, define their inference as an optimization process.
\end{description}

\subsection{Building Blocks}\label{sec:buildingblocks}

To carry out input purification or model adaptation, adaptive test-time defenses can rely on distinct building blocks.
We identify common blocks that span current defenses.
\begin{description}[leftmargin=0pt,itemsep=2pt,topsep=0pt,parsep=0pt]
\item[Iterative algorithm (IA):] All defenses in our case study use an iterative algorithm.
Five in nine solve their test-time optimization approximately by gradient descent, with some defenses using a single normalized gradient step \citep{qian2021improving} and others using as many as 40 gradient descent steps \citep{mao2021adversarial}.
Other defenses may also use layers defined via implicit functions whose output is computed with iterative algorithms (e.g., neural ODE; \citealp{kang2021stable}) or may rely on iterative generative models (e.g., energy-based models; \citealp{yoon2021adversarial}).
\item[Auxiliary networks (AN):] A significant proportion of adaptive test-time defenses (
seven in nine defenses studied) rely on an external network besides the underlying static model.
This network typically helps to compute the optimization objective (such as a self-supervised objective; \citealp{shi2020online}) or directly manipulates inputs (using a generative model for example; \citealp{yoon2021adversarial}).
It might be bound to \citep{mao2021adversarial,hwang2021aidpurifier} and possibly trained jointly with \citep{qian2021improving,shi2020online,chen2021towards} the static model, or used as a standalone component \citep{yoon2021adversarial}.
\item[Randomness (R):]
Three in nine defenses studied are randomized either explicitly (e.g., by adding noise to the input; \citealp{wu2021attacking}) or implicitly (e.g., by sampling different batches of inputs; \citealp{mao2021adversarial}).
As such, the same inputs may yield different outputs.
We exclude pure randomization defenses (without test-time optimization).
\item[External data (ED):] One of the defenses studied exploits a collection of additional images, which are combined with the inputs.
As such, inference does not depend only on the given input for classification \citep{mao2021adversarial}.
\end{description}

\subsection{Advantages, Drawbacks and Pitfalls}\label{sec:pitfalls}

Test-time adaptation by optimization has shown promising results for robustness to natural shifts like common corruptions \citep{sun2020test,wang2021tent}, and have the potential to likewise improve adversarial robustness.
Unlike static defenses, adaptive defenses have the freedom to alter an input (almost) arbitrarily or even update their own parameters on the input.
This additional complexity seems to offer an advantage over static defenses that rely on adversarial training but remain fixed during testing.
However, complex defenses are often difficult to evaluate \citep{tramer_adaptive_2020}, and adaptation might require techniques than those needed for static defenses.
The accurate evaluation of adaptive test-time defenses presents several challenges:
\begin{description}[leftmargin=0pt,itemsep=2pt,topsep=0pt,parsep=0pt]
\item[Obfuscated gradients:] First and foremost, iterative optimizers might be not differentiable (e.g., when using projections).
\gls{bpda}, which is typically used in such circumstances and usually implemented as an identity function for the backward pass, can be weaker for procedures consisting of many iterations.
Even in cases where the optimization is differentiable, gradients might vanish or explode when a large number of iterations is used.
\item[Randomness:] The use
of randomized elements requires to resort to methods like \gls{eot} \citep{athalye_obfuscated_2018} and make the task of the attacker more expensive.
\item[Runtime:] The cost of inference is significantly higher than for static models, up to hundreds of times, which clearly impacts the amount of effort required for an accurate evaluation.
Such high computational complexity calls into question the applicability of these defenses in the first place.
Regardless of running attacks for evaluation, classification in deployment can be extremely slow.
Hence, the trade-off in potential improvement versus additional computation should be the subject of further research.
\end{description}

\subsection{Threat Model and Adversarial Capabilities}\label{sec:threat}

We focus exclusively on the robustness of image classifiers to $\ell_p$-norm bounded input perturbations.
Given a $K$-way classifier $f:\R^d \rightarrow \R^K$, a test input $\vx \in [0, 1]^d$ with label $y$ and a norm bound $\epsilon > 0$, an attack succeeds if it finds a perturbation $\vdelta$ such that
\begin{align}
    &\argmax_{k \in \{1, \ldots, K\}} [f(\vx+\vdelta)]_k \neq y, \\
    &\textrm{with~} \norm{\vdelta}_p \leq \epsilon \textrm{~and~} \vx+\vdelta \in [0, 1]^d. \nonumber
\end{align}
where $[a]_i$ represents the $i$-th coordinate of $a$.
We assume that the attacker has full white-box access to the defense.
In other words, the attacker is aware of the test-time optimization in place, and has access to the parameters of the classifier and any auxiliary model(s).
For randomized defenses, the attacker does not have access to the state of the random number generator.
In some cases, we will fix the random seed to evaluate whether the impact of a proposed defense is mostly driven by optimization or randomization.
Finally, when the defense uses external data, the attacker may have access to the set of images used (when fixed and hard-coded by the defense) or may try to infer the distribution from which these external images are drawn.
For defenses that operate on a full batch of images, the attacker may be able to influence either one image from the batch or the whole batch.
For all defenses, we respect the threat model defined by each defense when clearly defined in the corresponding paper or code.

\section{Case Study of Adaptive Methods}\label{sec:casestudy}

\begin{table*}[t] \centering
\caption{%
Summary of the nine \emph{adaptive test-time defenses} evaluated in our case study.
We categorize each defense by its principles---input purification (IP) and model adaptation (MA)---and its building blocks: iterative algorithm (IA), auxiliary network (AN), randomization (R), and external data (ED).
We measure the robust accuracy of each defense (and in parenthesis that of its underlying static model) against $\ell_\infty$-norm bounded perturbations of size $\epsilon=8/255$ on \cifar (``Ours'') along with the robust accuracy measured by the respective papers (``Reported'').
* This evaluation uses $\epsilon=2/255$.
** This evaluation uses batch size 50 instead of the original 512.
}
\label{tab:summary_casestudy}
\vspace{2mm}
\resizebox{\textwidth}{!}{
\begin{tabular}{ll cc cccc c l rrl}
\multirow{2}{*}{Defense} & \multirow{2}{*}{Venue} & \multicolumn{2}{c}{Principles} & \multicolumn{4}{c}{Building blocks} & \multirow{2}{*}{\parbox{1.3cm}{\centering Inference time}} & \multirow{2}{*}{Evaluation method} & \multicolumn{3}{c}{Robust Accuracy} \\
\cmidrule(r){3-4} \cmidrule(r){5-8} \cmidrule(r){11-13}
& & IP & MA & IA & AN & R & ED & & & Reported & \multicolumn{2}{c}{Ours} \\
\toprule
\citet{kang2021stable}       & NeurIPS &        & \cmark & \cmark & \cmark &        &       &  2$\times$ & Transfer APGD               & 57.76\% & 52.2\% & (53.9\%) \\
\citet{chen2021towards}*      & ICLR &        & \cmark & \cmark & \cmark &        &        & 59$\times$ & APGD+BPDA                   & 34.5\% & 5.6\% & (0.0\%) \\
\citet{wu2021attacking}      & ArXiv & \cmark &        & \cmark &        & \cmark &        &  46$\times$ & Transfer APGD+BPDA+EoT      & 65.70\% & 61.0\% & (63.0\%) \\
\citet{alfarra2021combating} & AAAI & \cmark &        & \cmark &        &        &        & 8$\times$ & RayS (decision-based)       & 79.2\%  & 66.6\% & (66.6\%) \\
\citet{shi2020online}        & ICLR & \cmark &        & \cmark & \cmark &        &        & 518$\times$ & APGD+BPDA (traj.)           & 51.02\% & 3.7\% & (0.0\%) \\
\citet{qian2021improving}    & ArXiv & \cmark &        & \cmark & \cmark &        &        & 4$\times$ & APGD                        & 65.07\% & 12.6\% & (7.7\%) \\
\citet{hwang2021aidpurifier} & ICML(W) & \cmark &        & \cmark & \cmark &        &        & 40$\times$ & APGD+BPDA                   & 52.65\% & 43.8\% & (49.3\%)  \\
\citet{mao2021adversarial}**   & ICCV & \cmark &        & \cmark & \cmark & \cmark & \cmark & 407$\times$ & APGD+BPDA+EoT                    & 63.83\% & 58.4\% & (59.4\%)  \\
\citet{yoon2021adversarial}  & ICML & \cmark &        & \cmark & \cmark & \cmark &        & 176$\times$ & APGD+EoT                    & 69.71\% & 33.7\% & (0.0\%) \\
\bottomrule
\end{tabular}
}
\end{table*}

In our case study, we evaluate nine adaptive test-time defenses which rely on the adaptation principles elaborated in \autoref{sec:definition}.
One evaluation (of \citealp{qian2021improving}) is presented in the appendix due to space constraints.
Table~\ref{tab:summary_casestudy} summarizes the defenses considered in this case study, categorizes each defense (according to \autoref{sec:principles} and \autoref{sec:buildingblocks}), and details the corresponding results against $\ell_\infty$-norm bounded attacks with budget $\epsilon=8/255$ on \cifar (which is commonly evaluated by all defenses in their respective papers).\footnote{With the exception of $\epsilon=2/255$ for \citet{chen2021towards}.}
In particular, we report the robust accuracy (i.e., the classification accuracy on adversarially-perturbed inputs) originally reported in each work and the result of our evaluation.
In parentheses, we report the robust accuracy of the underlying static model.
We also measure the cost of inference computation in relation to performing the same inference with the underlying static model.
Note that many factors can influence this value (e.g., architecture of the classifier, compute infrastructure, optimized implementation) and we use the runtime observed in our evaluation.

Overall, we find that the reported accuracy is consistently overestimated in all papers.
Five defenses \citep{yoon2021adversarial,hwang2021aidpurifier,qian2021improving,shi2020online,chen2021towards} have robust accuracies well below 50\% and are not competitive with state-of-the-art static models (even of moderate size; \citealp{gowal2021improving,rade2021helperbased}).
Most importantly, four defenses \citep{wu2021attacking,kang2021stable,hwang2021aidpurifier,mao2021adversarial} weaken the underlying static defense (when it is already robust), while the others provide minor improvements at major computational cost.

Alongside our case study, we index additional adaptive test-time defenses in the appendix.
In contemporary work, \citet{chen2022towards} carry out a complementary analysis of transductive test-time defenses~\citep{wu2020rmc,wang2021fighting}, which depend on multiple test inputs and even joint optimization across training and testing data.

\subsection{Evaluation Methods}

For completeness, we give a short overview of the attacks and techniques used in our case study.
More details are described in the evaluation of each defense.
{\bf \autoattack} \citep{croce_reliable_2020} is a common benchmark for evaluating static defenses, but it is not designed for adaptive test-time defenses.
{\bf \gls{apgd}} \citep{croce_reliable_2020} is a variant of \gls{pgd}---one of the most popular technique for $\ell_p$-norm bounded adversarial attacks. Its varied surrogate losses include cross-entropy, Carlini-Wagner (CW), margin \citep{carlini_adversarial_2017}, or the (targeted) \gls{dlr} \citep{croce_reliable_2020}.
{\bf RayS} \citep{chen_rays_2020} is a decision-based attack designed for $\ell_\infty$-norm bounded perturbations. It only requires the label predicted by the model.
{\bf \gls{bpda}} \citep{athalye_obfuscated_2018} permits the attack of non-differentiable defenses by approximating them with differentiable functions during gradient computation. The identity is a common approximation.
{\bf \gls{eot}} \citep{athalye_obfuscated_2018} permits the attack of randomized defenses. The predictions and gradients are computed in expectation over the randomness of the model, approximated by averaging the results of multiple runs with the same input.
{\bf Transfer attacks} generate adversarial perturbations on a surrogate model and use them on the target model.

\subsection{Stable Neural ODE with Lyapunov-Stable Equilibrium Points for Defending against Adversarial Attacks \citep{kang2021stable}} \label{sec:kang2021stable}

\paragraph{Summary of method.}

This defense applies a neural \gls{ode} layer to improve robustness against $\ell_p$-norm bounded attacks by the stability of equilibrium points of the underlying \gls{ode}.
The proposed \sodef model consists of a feature extractor followed by a neural ODE layer and finally one or multiple fully connected layers.
The feature extractor is a standard convolutional network, while the neural ODE can be solved with numerical methods (Runge-Kutta of order 5 in this case) up to some integration time $T$.

\paragraph{Evaluation.}

While \citeauthor{kang2021stable} consider many scenarios in their experimental evaluation, we test \sodef when the feature extractor consists of a robust classifier, trained with \trades \citep{zhang_theoretically_2019} against $\ell_\infty$-norm perturbations of size $8/255$ on \cifar, without the last linear layer (since checkpoints are available online).
We use the original implementation with corresponding parameters, including integration time $T=5$.
A significant improvement in robustness against $\ell_\infty$- and $\ell_2$-norm perturbations is reported (57.76\% and 67.75\%) when compared to the original \trades model (53.69\% and 59.42\%), under the evaluation of \autoattack.
In particular, \autoattack is used both directly on the \sodef model and as the basis for a transfer attack from the \trades classifier.
However, we note that by default \autoattack returns the original images when they are originally misclassified or when no adversarial perturbation is found.
This means that, in practice, most of the transferred points are clean images (since the \trades model is highly robust).
Moreover, \autoattack does not aim at maximizing the confidence in the misclassification after this is achieved and, as such, it might be not the strongest method for transfer attacks.
As a consequence, we use \gls{apgd} to maximize different target losses on all clean inputs using the \trades classifier and then test them on \sodef.
We run \gls{apgd} for 100 iterations with cross-entropy, \gls{cw} and targeted \gls{dlr} (with 3 target classes) loss, and consider the transfer attack successful if any of the resulting points fools the \sodef model.
Table~\ref{tab:sodef} shows the results of the described experiments for $\ell_\infty$- and $\ell_2$-norm bounded attacks (on 1000 test images with $\epsilon=8/255$ and $\epsilon_2=0.5$, respectively).
When transferring inputs that maximize the surrogate losses, the robustness of \sodef is lower than that of the original \trades classifiers.

\begin{table}[t]
    \centering\small
    \caption{Robust accuracy on 1000 points against $\ell_p$-norm bounded attacks: \trades is tested with \autoattack, and \sodef with transferring APGD from the static \trades model.}
    \label{tab:sodef}
    \vspace{2mm}
    \begin{tabular}{c c cc}
         \multirow{2}{*}{Threat model} & Static & \multicolumn{2}{c}{with \sodef} \\
         \cline{3-4}
         & AA & AA (transf.) & APGD (transf.) \\
         \toprule
         $\ell_\infty$ & 53.9\% & 60.9\% & 52.2\% \\
         $\ell_2$ & 59.1\% & 66.8\% & 58.2\% \\
         \bottomrule
    \end{tabular}
    \vspace{-2mm}
\end{table}

\subsection{Towards Robust Neural Networks via Close-loop Control \citep{chen2021towards}}

\paragraph{Summary of method.}

Using the assumption that the representation of an adversarial input becomes progressively more corrupted as it moves through the layers of a deep network, this defense corrects its trajectory by adding offsets to intermediate activations of a pre-trained static model.
At each layer $i$, \emph{control parameters} $\vu_i$ are added to its output.
These control parameters are optimized to reduce the reconstruction error between the original activations and corresponding ``reduced'' activations.
These reduced activations are encoded onto a lower-dimensional manifold and decoded back to the full state space using \gls{pca} or shallow auto-encoders.
At inference time, the parameters $\vu_i$ are jointly optimized with multiple steps of gradient descent and then used to produce the final prediction.

\paragraph{Evaluation.}

\citeauthor{chen2021towards} report 11\% robust accuracy against \gls{pgd} attacks of size $\epsilon=8/255$.
This is not competitive with state-of-art robust classifiers and already hints that the defense may be ineffective.
As such, we focus on their strongest result, at the smaller $\epsilon=2/255$, where they report 50\% robust accuracy.
Even this reduced claim is voided by further evaluation.
Crucially, their attack is not adaptive, as it does not exploit knowledge of the defense.
\citeauthor{chen2021towards} argue that adaptive attacks are infeasible, as they would need to steer the control parameters $\vu_i$ toward zero.
They also argue that calculating the gradients of the optimization process is too difficult.
Neither argument is sound, as the defense might still fail for nonzero values of $\vu_i$, and obfuscated gradients can be approximated.
We demonstrate this by combining \gls{pgd} and \gls{bpda}.

As pre-trained models are not publicly available, we train our own ResNet-20 model, and learn the linear embeddings with \gls{pca} using the code available.
Unfortunately, some hyperparameters are not documented, and so we find a working configuration (using 5 iterations and a learning rate of $5\cdot10^{-3}$) by grid search that approximately reproduces the results of the paper.
The resulting defense obtains 89.1\% accuracy on clean images.
We also reproduce their 20-step \gls{pgd} evaluation by attacking the underlying classifier and testing the adversarial examples against their defense, and measure a robust accuracy of 34.5\% (compared with 0\% for the static model).
The robustness gain is slightly less than reported (i.e., 50\%) but still enough for our demonstration.\footnote{This is the only configuration matching the reported accuracy on clean images while having non-trivial robustness against PGD.}
Each prediction takes 4 backward passes and 5 forward passes.

We now mount an adaptive attack.
At test-time, the defense runs multiple iterations of gradient descent to optimize the control parameters $\vu_i$.
The optimized $\vu_i^*$ are then used for the final prediction.
Since $\vu^*_i$ is merely added to the activations of the $i$-th layer, it is possible to derive an adversarial gradient for this final forward pass.
Although this can only approximate the gradient of the full defense, we find that it suffices to drastically improve the attack success rate.
We use this gradient with \gls{bpda} and repeat the same 20-step PGD attack.
The dynamic model now achieves only 9.5\% robust accuracy (instead of 34.5\%).
Finally, the gradient approximation also enables attack by \gls{apgd} with \gls{bpda}, which drops the robust accuracy further to 5.6\%.

\subsection{Attacking Adversarial Attacks as a Defense \\ \citep{wu2021attacking}}
\label{sec:wu2021attacking}

\begin{table*}[t]
    \centering \small
    \caption{Robust accuracy on 1000 points of various static models (measured by \gls{apgd}) and their adaptive version with \gls{hd} (using 20 steps), where we transfer \gls{apgd} attacks obtained on the static model and \gls{apgd} with \gls{bpda} and \gls{eot} attacks from the model using \gls{hd} with 5 steps. We also include the worst-case among both transfer attacks.}
    \label{tab:hedge-def}
    \vspace{2mm}
    \begin{tabular}{l c ccc}
         \multirow{2}{*}{Underlying model} & Static & \multicolumn{3}{c}{Transfer to \gls{hd} with 20 steps using as surrogate} \\
         \cline{3-5}
         & defense & Static & \gls{hd} (5 steps) & Worst-case \\
         \toprule
         \citet{gowal_uncovering_2020} & 63.0\% & 65.9\% $\pm$ 0.09 & 61.1\% $\pm$ 0.14 & 61.0\% $\pm$ 0.14 \\
         \citet{carmon_unlabeled_2019} & 59.2\% & 65.5\% $\pm$ 0.34&59.6\% $\pm$ 0.22 & 59.2\% $\pm$ 0.32 \\
         \citet{andriushchenko_understanding_2020} & 45.0\% &54.1\% $\pm$ 0.15 & 45.8\% $\pm$ 0.12 & 45.2\% $\pm$ 0.28 \\
         \bottomrule
    \end{tabular}
    \vspace{-2mm}
\end{table*}

\paragraph{Summary of method.}

The \gls{hd} proposed by \citeauthor{wu2021attacking} aims to defend against adversarial perturbations by \emph{maximizing} the cross-entropy loss of the classifier's predictions summed over all classes.
The intuition is that the images that are wrongly classified have stronger gradients, which dominate the optimization and easily reduce the confidence in the initial decision, while correctly classified points are minimally impacted by the defense.
The proposed method does not require modifications of the training scheme of the underlying static model.
\gls{hd} solves, for a classifier $f$,
\begin{align}
&\argmax_{\vdelta} \, \sum_{k=1}^K L_\textrm{ce}(f(\vx + \vdelta), k), \\
&\textrm{with} \; \norm{\vdelta}_\infty \leq \epsilon_\textrm{d}, \; \vx + \vdelta \in [0, 1]^d  \nonumber
\end{align}
with $20$ steps of PGD (and random initialization), $\epsilon_\textrm{d}=8/255$ (the same used by the attacker) and step size $\eta=4/255$.
This preprocesses the input to the classifier $f$, and therefore applies to any model.
Inference with \gls{hd} costs 21 forward and 20 backward passes of the model.

\paragraph{Evaluation.}

The main experimental evaluation in \citet{wu2021attacking} consists in transferring perturbations generated by several attacks on the underlying classifier $f$ to the model equipped with \gls{hd}.
This results in absolute improvements of 2\% to 4\% in robust accuracy.
An adaptive attack, equivalent to \gls{bpda}, is developed on the full defense, but it does not consider the randomness of the defense.
This attack is reported to slightly reduce the effectiveness of \gls{hd} on one model.
The official code is not provided.
Hence, we implement the defense algorithm ourselves using the details available in the paper.

\citeauthor{wu2021attacking} report that \gls{pgd} is often more effective than stronger methods like \autoattack.
We hypothesize that this is caused by the original implementation of \autoattack which returns copies of unperturbed inputs when the attack is unsuccessful or when the original input is already misclassified.
However, \gls{hd} might also turn an originally correctly classified input into a misclassified one, and this is more likely if it is close to the decision boundary (as we expect an unsuccessful adversarial example to be).
Ultimately, we use \gls{apgd} on the targeted \gls{dlr} loss with 5 target classes (that is 5 restarts of 50 iterations).
We use \gls{bpda} and reduce the number of steps of \gls{hd} to 5 when crafting the perturbations.
To counter the randomness of the initial step of \gls{hd}, we use 4 steps of \gls{eot}.
We evaluate the obtained perturbations on the full defense with 20 steps.
In Table~\ref{tab:hedge-def}, we report the robust accuracy obtained with our evaluation on static models available in \robustbench~\citep{croce2020robustbench}.
We also report the robust accuracy obtained by transferring adversarial perturbations from the underlying static models.
We observe that \gls{hd}
does not provide clear improvements to the robustness of static models, and in one case it weakens it.

\subsection{Online Adversarial Purification based on Self-Supervision \citep{shi2020online}}
\label{sec:shi2020online}

\paragraph{Summary of method.}

This defense purifies adversarial perturbations by minimizing an auxiliary loss before performing inference.
The auxiliary loss is connected to a self-supervised task, which may require an additional network which shares some feature representation with the classifier.
Since the auxiliary loss is also optimized at training time, the system should learn to perform classification based on robust features that are shared across the supervised and auxiliary tasks.
In the evaluation of \citet{shi2020online}, \gls{fgsm} is often a stronger attack than \gls{pgd}, which contradicts the principles of \citet{carlini_evaluating_2019}.

We focus on the defense variant that uses label consistency as auxiliary task since it does not require an additional network at test-time and yields the best results on \cifar.
It uses as auxiliary loss
\begin{align}
    L_\textrm{aux}(f, \vx) = \norm{f(a_1(\vx)) - f(a_2(\vx))}_2,
\end{align}
with $f$ the classifier and $a_1, a_2$ augmentations of the input.
At test-time, the problem
\begin{align}
&\argmin_{\vdelta} \, L_\textrm{aux}(f, \vx + \vdelta), \\
&\textrm{with} \; \norm{\vdelta}_\infty \leq \epsilon_\textrm{d}, \; \vx + \vdelta \in [0, 1]^d \nonumber
\end{align}
is optimized with 5 steps of \gls{pgd} (without random initialization).
The procedure is repeated for eleven values of $\epsilon_\textrm{d}$ and that attaining the lowest loss is chosen.
Overall, inference requires $5 \cdot 11 \cdot 2$ forward and backward passes during purification plus $1$ forward pass for the final classification.

\paragraph{Evaluation.}

We consider the small pre-trained model (referred to as ResNet-18 in \citealp{shi2020online}).
It achieves a clean accuracy of 83.7\% (after purification) on the first 1000 test images of \cifar.
The original evaluation reports for this model a robust accuracy of 51.02\% under transfer from an \gls{fgsm} attack on the static model.
We run \gls{apgd} with \gls{bpda} on the cross-entropy loss, but, instead of using as update direction (the sign of) the gradient of $f$ with respect to the final purified image only (as one would get by approximating the whole purification process with the identity function), we average gradients over intermediate iterates produced by the purification process.
Intuitively, this steers all intermediate images towards misclassification and make the attack more effective.
Running this attack with 1000 iterations reduces the robust accuracy of the defense to 3.7\%.

\subsection{Adversarial Attacks are Reversible with Natural Supervision \citep{mao2021adversarial}}

\paragraph{Summary of method.}

This defense states that images contain intrinsic structure that enables the reversal of adversarial attacks.
In particular, they modify the inference step to purify the input using a trained contrastive representation $g$ that leverages the intermediate activations of a pre-trained static model $f$.
Given an input $\vx$, the defense creates a modified input $\vx + \vdelta^\star$ where $\vdelta^\star$ is the solution to
\begin{align}
&\argmin_{\vdelta} \, L_\textrm{aux}(g(t_1(\vx + \vdelta)), g(t_2(\vx + \vdelta))), \\
&\textrm{with} \; \norm{\vdelta}_\infty \leq 2\epsilon, \; \vx + \vdelta \in [0, 1]^d, \nonumber
\end{align}
and is found using $N$ \gls{pgd} steps.
$L_\textrm{aux}$ is the contrastive loss and $t_1$, $t_2$ are two transformations randomly sampled from a set of predefined image augmentations.
The authors use either two or four augmentations for each image and leverage a separate set of images to build negative pairs (to be used in the contrastive loss).
The paper remains unclear about the provenance of these additional images.
The released code uses augmented views of other attacked images to build the negative pairs and thus operates at the level of a batch, which can both weaken the defense and attack.
At inference, the modified input is fed to a static model.
In practice, \citeauthor{mao2021adversarial} set $N=40$.

\paragraph{Evaluation.}

According to \citeauthor{mao2021adversarial}'s evaluation their best model achieves a robust accuracy of 67.79\% against \autoattack, 64.64\% against \gls{pgd} with 50 steps and 63.83\% against the Carlini-Wagner attack with 200 steps~\citep{carlini_towards_2017} on \cifar against $\ell_\infty$-norm bounded perturbations of size $\epsilon=8/255$.
These attacks are performed on the underlying static model and transferred to the full defense.
At first sight, it can seem surprising that \autoattack is weaker since it consists of a suite of attacks with numerous restarts and many steps.
However, by default \autoattack returns the original images when these are originally misclassified or when no adversarial perturbation is found.
This means that, in practice, most of the transferred points are clean images (since the underlying static model is already robust).

We replicate the setup from the authors which operates at the batch level.
We use four image transformations and set the number of iterations to 40.
For the purpose of this demonstration, and to be able to run our evaluation on a single NVIDIA V100 GPU, we reduce the batch size to 50 (instead of 512).
While this change may negatively impact the defense, we found that transfer attacks match the results from \citet{mao2021adversarial}: Using \autoattack from the static model yields a robust accuracy of 67.0\% (compared to 67.79\%); Using a custom implementation of \gls{apgd} (which returns worst-case adversarial examples) on the cross-entropy and \gls{dlr} losses with 10 steps and 10 restarts, we obtain a robust accuracy of 63.9\%, which in line with the worst-case robust accuracy obtained by \citeauthor{mao2021adversarial} (i.e., 63.83\%).
We note that under the same \gls{apgd} attack, the static model obtains a robust accuracy of 59.4\%.
When attacking the full adaptive defense, we use \gls{bpda} and 16 \gls{eot} iterations.
We obtain a robust accuracy of 58.4\% and conclude that the proposed defense weakens the underlying static model.
To determine the effect of randomness, we also perform our evaluation by fixing the random seed (and removing \gls{eot}).
Without randomness, we obtain a robust accuracy of 56.4\%.

\subsection{\aidpurifier: A Light Auxiliary Network for Boosting Adversarial Defense \citep{hwang2021aidpurifier}}

\paragraph{Summary of method.}

This defense, proposed by \citet{hwang2021aidpurifier}, uses a discriminator to purify the input of a pre-trained classifier.
The discriminator is trained to distinguish adversarially perturbed from clean inputs.
It exploits the intermediate activations of the underlying static model.
At inference time, every input is purified by minimizing the probability of perturbation according to the discriminator, with the goal of reducing any adversarial effect the input may have.

Given a discriminator $g$, an input $\vx$, $\epsilon_\textrm{d}> 0$, \aidpurifier approximately solves the problem
\begin{align}
&\vdelta^\star \approx \argmin_{\vdelta} \, g(\vx + \vdelta) \\
&\textrm{with} \; \norm{\vdelta}_\infty \leq \epsilon_\textrm{d}, \; \vx + \vdelta \in [0, 1]^d  \nonumber
\end{align}
with $N$ steps of \gls{pgd} and a fixed step-size $\alpha$ (the values of $N$ and $\alpha$ vary across datasets).
Inference is performed using the pre-trained classifier on the point $\vx + \vdelta^\star$.
The inference costs $N$ additional forward and backward passes of the discriminator compared to the standard one (in practice $T=10$ is used).

\paragraph{Evaluation.}

The original evaluation relies mostly on transferring perturbations adversarial to the pre-trained classifier to the classifier endowed with purification.
The authors also propose an adaptive attack which optimizes a convex combination of the classification loss and the output of the discriminator.
This adaptive attack is not effective in decreasing the robust accuracy compared to the non-adaptive counterpart.
We focus our analysis on static models trained using adversarial training \citep{madry_towards_2017} as their respective pre-trained discriminators are available publicly.
We note that the reported improvement of \aidpurifier on these models is rather small (1 to 2\%) for all datasets, with the exception of \svhn where clean and robust accuracy increase by 22\% and 27\%, respectively.
In the following, we consider the $\ell_\infty$-norm bounded perturbations of size $\epsilon=8/255$ for \cifar and \cifarh, and of size $\epsilon=12/255$ for \svhn.
We evaluate the robustness of the static models with \autoattack, and that of the dynamic models with \gls{apgd} combined with \gls{bpda}.
Table~\ref{tab:aid-purifier} shows that for \cifar and \cifarh the dynamic classifiers have lower robustness than the original static models.
While \aidpurifier seems to improve robustness on \svhn, we notice that, unlike on the other datasets, the static model has a low robust accuracy.\footnote{For reference, we can obtain a classifier with more than 40\% robust and 86\% clean accuracy by standard adversarial training.}

\begin{table}[t]
    \centering\small
    \caption{Robust accuracy on 1000 points against $\ell_\infty$-norm bounded attacks of static models alone and with \aidpurifier.} \label{tab:aid-purifier}
    \vspace{2mm}
    \begin{tabular}{l c c cc}
         \multirow{2}{*}{Dataset} & \multirow{2}{*}{$\epsilon$} &  \multirow{2}{*}{Static model} & \multicolumn{2}{c}{\aidpurified model} \\
         \cline{4-5}
         & & & Transfer & Direct \\
         \toprule
         \cifar & 8/255 & 49.3\% &56.0\% & 43.8\% \\
         \cifarh & 8/255 & 23.3\% & 32.3\%& 18.2\% \\
         \svhn & 12/255 & 10.5\% & 62.0\% & 29.1\% \\
         \bottomrule
    \end{tabular}
    \vspace{-2mm}
\end{table}

\subsection{Combating Adversaries with Anti-Adversaries \citep{alfarra2021combating}}

\paragraph{Summary of method.}

This defense tries to counter gradient-based based attacks during inference by maximizing the classifier's confidence.
It uses a static model $f$ to predict pseudo-labels $\hat{y} = \argmax_k [f(\vx)]_k$ for each input $\vx$, and optimizes $\vx' = \argmin_{\vx'} L_\textrm{ce}(f(\vx', \hat{y}))$ with $N$ gradient descent steps to increase the classifier's confidence in $\hat{y}$ (with $L_\textrm{ce}$ being the cross-entropy loss).
The goal is to cancel out attacks that aim to decrease this confidence score.
In practice, $N$ is set to $2$, which leads to an increase in inference time of $8\times$ (the original implementation also makes an additional unnecessary forward pass).

This design implies that the defense inherits the decision boundary of the underlying classifier, as increasing a classifier's confidence in its own predictions will not change its decision.\footnote{We observe that the decision boundary can sometimes change due to numerical inaccuracies. This effect is rare and disappears when reducing the defense step-size.}
Consequently, there is no fundamental gain in robustness and any measured increase can only be the result of obfuscation: forcing the classifier to report high confidence almost everywhere flattens the loss landscape of the cross-entropy loss and causes numerical problems not only for \gls{pgd}, but also for any score-based attack such as the Square attack~\citep{andriushchenko_square_2019}.

\paragraph{Evaluation.}

We focus on the strongest result reported by \citeauthor{alfarra2021combating} which consists of applying their defense for two iterations to a robust static \cifar model pre-trained with \gls{awp} \cite{wu_awp_2020}.
As noted above and as noted by \citeauthor{alfarra2021combating}, the defense can easily be circumvented by transferring adversarial examples from the underlying static model.
Using \gls{apgd} on the cross-entropy loss, we can reduce the robust accuracy of both the static and adaptive model to 63.7\% against $\ell_\infty$-norm bounded perturbations of size $\epsilon=8/255$ on \cifar.

However, we find that the defense is ineffective even against black-box decision-based attacks that require no knowledge of the static model at all.
Testing the defense against a range of attacks (which exclude transfer attacks), \citeauthor{alfarra2021combating} report a robust accuracy of 79.21\%.
They also conduct an evaluation against decision-based attacks with inconclusive results as they report a robust accuracy of 86.0\% against both the adaptive and static model, which is close the clean baseline of 88.0\%.
This suggests an incorrect use of the attack, or perhaps not enough iterations.
As a consequence, we apply the more efficient RayS attack with 10K queries.
We obtain a robust accuracy of 66.6\%, which is much lower than the result reported by \citeauthor{alfarra2021combating} (i.e., 79.2\%).
We also obtain 66.6\% against the static model, which shows that the defense has no effect against an attack that does not use confidence scores.

\subsection{Adversarial Purification with Score-based Generative Models \citep{yoon2021adversarial}}

\paragraph{Summary of method.}

This defense preprocesses the input of a classifier with an \gls{ebm} trained, with \gls{dsm}, to learn a score function to denoise perturbed images.
A particularity is that they allow only a few deterministic updates in the purification process (after adding noise to the initial input).

The proposed defense, named \gls{adp}, purifies an input $\vx$ with the iterative scheme, for $i=1, \ldots, T$.
\begin{align}
    \vx_0 = \vx + \vxi, \quad \vx_i = \vx_{i-1} + \alpha_{i-i} s(\vx_{i-1})
\end{align}
with $\vxi \sim \mathcal{N}(0, \sigma^2 I)$ an initial noise addition, $\alpha_i$ an adaptive step-size, and $s$ the score function approximated by the \gls{ebm}.
The procedure is repeated $S$ times, getting $S$ points $\vx_T^{(1)}, \ldots, \vx_T^{(S)}$.
Given a classifier $f$, the final classification is obtained as
\begin{align}
    \argmax_{k=1, \ldots, K} \frac{1}{S}\sum_{s=1}^S [\softmax(f(x_T^{(s)})]_k.
\end{align}
In practice, \citeauthor{yoon2021adversarial} set $T=10$, $S=10$ and $\sigma=0.25$, and $f$ is a pre-trained classifier.
Hence, the inference costs $S$ forward passes of $f$ and $2 \cdot T\cdot S$ forward passes of $s$ (since an additional call of the score networks is necessary for the computation of $\alpha_i$).

\paragraph{Evaluation.}

In their experimental evaluation, \citet{yoon2021adversarial} report that \gls{adp} achieves $69.71\%$ against $\ell_\infty$-norm bounded perturbations of size $\epsilon=8/255$ on \cifar.
The strongest attack is reported to be \gls{pgd} with \gls{bpda} and \gls{eot}, where \gls{bpda} approximates the iterative purification process with an identity.
However, we note \gls{bpda} is not necessary since the defense consists only in forward passes of the score network and of the classifier from which gradients can be computed.
As such, we apply 10 steps of \gls{apgd} combined with 200 \gls{eot} iterations to alleviate the randomness induced by $\vxi$.
This attack yields a robust accuracy of $33.7\% \pm 0.54$ averaged over 1000 images.\footnote{Average and standard deviation are computed over 5 repeated evaluations over \gls{adp} of the adversarially perturbed points.}

\section{Discussion and Recommendations}

Our case study shows that evaluating adaptive test-time defenses is more challenging than evaluating static defenses because of several complicating factors.
These factors include the possibly intricate optimization process, randomness, and the large computational requirements of running these defenses.
Moreover, each defense has its own peculiarities, which frustrates attempts at standardizing evaluation.
In fact, Table~\ref{tab:summary_casestudy} shows that the most effective attack differs across cases.
At the same time, the main elements constituting such attacks consist of techniques proposed in prior works, which can be adaptively combined depending on the elements of the defense.
The main challenge is therefore to find the attack setup which is the most suitable for each case.

\citet{carlini_evaluating_2019} proposed guidelines to evaluate adversarial robustness, and applied their principles to build a strong attack methodology.
Moreover, they identify signs of overestimated robustness (e.g., single-step attacks stronger than multi-step ones), which also hold for adaptive test-time defenses.
Our case study suggests taking the following steps to extend their foundational guidance:
\begin{itemize}[leftmargin=0.4cm,noitemsep,topsep=0pt]
\item[\ding{182}] Transfer attacks that are effective against the underlying static model.
Consider using other models as surrogates, such as those available in the \robustbench model zoo.
When transferring attacks, make sure to transfer unsuccessful adversarial attacks (i.e., perturbations that attain high loss but do not lead to misclassification) rather than transferring the unperturbed input (as a library may do by default, e.g., \autoattack).
\item[\ding{183}] Verify with various black-box attacks that the adaptive test-time defense is more robust than the underlying static model. Consider Square \citep{andriushchenko_square_2019} which is score-based and RayS \citep{chen_rays_2020} which is decision-based for this purpose.
\item[\ding{184}] Apply strong white-box attacks to the full defense when possible. We found \gls{apgd} (with multiple losses and restarts) to be a reliable gradient-based attack in most cases.
  Modern frameworks such as \texttt{PyTorch} \citep{pytorch} and \texttt{JAX} \citep{bradbury_jax_2018} enable easy gradient computation.
\item[\ding{185}] When gradient attacks fail, due to non-differentiability or vanishing gradients caused by excessive iteration, consider combining \gls{apgd} with \gls{bpda} and try various approximations for the backward pass.
  In particular, \gls{bpda} can simply replace the backward pass for the optimization process with the identity, or make use of intermediate iterates (see \autoref{sec:shi2020online}).
\item[\ding{186}] When randomness is present, explictly or implictly, use \gls{eot}, or remove randomness altogether by fixing the seed at each attack step.
\item[\ding{187}] Ultimately, although they can serve as baselines, attacks developed for static defenses are not guaranteed to be effective for adaptive test-time defenses.
Consequently, always try to implement adaptive attacks that are specific to adaptive defenses.
These attacks should be stronger than their non-adaptive counterparts, unlike the results reported in \citet{mao2021adversarial} and \citet{hwang2021aidpurifier}.
\end{itemize}

\section{Conclusion}

Adaptive test-time defenses complicate robustness evaluation due to their complexity and computational cost.
Despite these complications, our evaluation succeeds in reducing the apparent robustness of the studied defenses, with a relative reduction of more than $50\%$ for four of the nine defenses.
In all cases with an adversarially robust static model (five out of nine), the adaptive test-time defense does not improve upon it and might even weaken it.
These results are disappointing, but we foresee that adaptive test-time defenses can still potentially lead to significant robustness gains, and clearly our results do not challenge the whole idea.
While we could not provide a standardized evaluation protocol for adaptive test-time defenses, we hope that our case study and recommendations can guide future evaluations as new defenses are developed.
Furthermore, we emphasize the need to measure gains in robustness against the computation required.
This weakness may be turned into a potential strength, if the additional computation used by the defender can impose even more computation on the attacker.
Finally, we note that the code developed for our case study is available at \url{https://github.com/fra31/evaluating-adaptive-test-time-defenses}.

\section*{Acknowledgements}

Francesco Croce and Matthias Hein acknowledge support from DFG grant 389792660 as part of TRR 248.
We would like to thank the authors of the papers evaluated in our case study, who have graciously answered our questions.
We also thank Andrei Rusu for participating in our initial discussions and for helping review this manuscript.

\bibliography{bibliography}
\bibliographystyle{icml2022}

\clearpage

\appendix

\section{Case Study of Adaptive Test-time Defenses}

We evaluate the remaining defense included in our case study, which is only reported here due to space constraints.

\subsection{Improving Model Robustness with Latent Distribution Locally and Globally \citep{qian2021improving}}

\paragraph{Summary of method.}

This defense introduces a new technique called \gls{atld}.
\gls{atld} builds adversarial examples by fooling a discriminator network $g$.
The discrimator has $C + 1$ outputs, where the zero-th output distinguishes between clean and adversarial images, while the remaining outputs predict the class.
In a process similar to \citet{madry_towards_2017}, adversarial examples are used within a cross-entropy loss to train a static model.
The claimed benefit of this approach is that adversarial examples are not biased towards the decision boundary of the classifier.
The experiments demonstrate that the resulting model (denoted \gls{atld}-) is more robust to weak adversarial attacks such as \gls{pgd} with 10 steps.
In addition to \gls{atld}, the authors also propose a modification to the inference process called \gls{imf}.
They modify inference by repairing inputs $\vx$ to $\vx + \vdelta^\star$ such that the trained discrimator is more likely to classify them as clean examples.
The optimal purification offset $\vdelta^\star$ is found using \gls{pgd} with $N$ steps to solve the problem
\begin{align}
&\argmin_{\vdelta} \, L_\textrm{softmax} [g(\vx + \vdelta)]_0, \\
&\textrm{with} \; \norm{\vdelta}_\infty \leq \epsilon_\textrm{d}, \; \vx + \vdelta \in [0, 1]^d.  \nonumber
\end{align}
The resulting models, dubbed \gls{atld} and \gls{atld}+, are seemingly more robust to evaluation pipelines like \autoattack (reaching 65.07\% robust accuracy against $\ell_\infty$-norm bounded perturbations of size $\epsilon=8/255$).
In practice, \citeauthor{qian2021improving} set $N=1$ and $\epsilon_\textrm{d}=2\epsilon$.

\paragraph{Evaluation.}

As noted by \citeauthor{qian2021improving}, the \gls{atld}- model is only robust to weak attacks and completely breaks under \autoattack.
In our evaluation, using \gls{apgd} on cross-entropy and targeted \gls{dlr} losses (with 9 targets), the robust accuracy of the \gls{atld}- model reduces to 7.7\% (compared to 65.40\% against \gls{cw} with 100 steps).
The authors evaluate their full defense, which includes \gls{imf}, by transferring attacks from various static models.
Surprisingly, transferring from the underlying \gls{atld}- model results in weaker attacks than transferring from another model trained through adversarial training~\citep{madry_towards_2017}.
This suggests that the underlying static model obfuscates gradients.
Consequently, we increase the number of restarts to 100.
Since the defense consists of a single differentiable step, we attack the full defense with \gls{apgd} on the cross-entropy and targeted \gls{dlr} losses.
Overall, we obtain a robust accuracy of 12.6\% for the \gls{atld}+ model, which is significantly lower than reported (i.e., 65.07\%).

\section{Additional Evaluation Details}\label{sec:app_casestudy}

We further detail the evaluation of the defenses described in our case study.
Please note that the code for each evaluation is available for clarity and reproducibility.

\subsection{Stable Neural ODE with Lyapunov-Stable Equilibrium Points for Defending against Adversarial Attacks \citep{kang2021stable}}

We use the original implementation of the defense and the corresponding checkpoints which are publicly available.\footnote{\url{https://github.com/kangqiyu/sodef}}
In the main manuscript (\autoref{sec:kang2021stable}), we use a WideResNet-34-10 as the underlying architecture for the static model.
As a point of comparison, we also evaluate \sodef when combined with a larger WideResNet-70-16.
The base model from \citet{rebuffi2021fixing} achieves a robust accuracy of 66.56\%.
According to \citet{kang2021stable}, this model combined with \sodef achieves 71.28\% when evaluated with \autoattack.
Similarly to what is described in \autoref{sec:kang2021stable}, we use \gls{apgd} to maximize different loss functions (cross-entropy, margin and targeted \gls{dlr} with 9 restarts) on the base model, and transfer the obtained perturbations to the \sodef model.
This yields a robust accuracy of 65.02\%, which can be further decreased to 64.20\% by taking the worst-case between the transfer attack and \autoattack.

\subsection{Towards Robust Neural Networks via Close-loop Control \citep{chen2021towards}}

We use the official implementation of the defense, which is publicly available.\footnote{\url{https://github.com/zhuotongchen/Towards-Robust-Neural-Networks-via-Close-loop-Control}} At the time of our evaluation, some hyperparameter choices in the code deviated from those in the paper while others were included in neither. Hoping to reproduce the results in the paper, we evaluate the following configuration:
\begin{itemize}[leftmargin=0.4cm,noitemsep,topsep=0pt]
    \item We use 5000 images from the \cifar training set to learn the linear projection.
    \item We set the regularization strength (decaying the magnitude of control parameters) to zero, disabling regularization. This matches the published code. We do not observe any notable effect from using regularization.
    \item We set the number of defense iterations to $N=5$ and learning rate to $\alpha=10^{-3}$, which we find to be the only configuration that is close to the robustness results claimed in the paper, while at the same time retaining good accuracy on clean images.
    \item We apply the defense to a ResNet-20 trained with the provided code. As no pre-trained checkpoints are available, we refrain from performing costly adversarial training and evaluate only on this normally-trained model.
\end{itemize}

For the attack, we first run \gls{pgd} with 20 steps of size $\frac{\epsilon}{4}$, mirroring the evaluation by \citet{chen2021towards} but adding \gls{bpda}.
We then run \gls{apgd} with \gls{bpda} using the cross-entropy loss for 100 iterations (no restarts).

\subsection{Attacking Adversarial Attacks as A Defense \citep{wu2021attacking}}

The official code for \gls{hd} is not available.
As such, we implement it by following the details provided by \citet{wu2021attacking}.
The models used in our evaluation are the WideResNet-28-10 trained with extra data \citep{gowal_uncovering_2020, carmon_unlabeled_2019} and the Pre-Activation ResNet-18 \citep{andriushchenko_understanding_2020}.
As stated in \autoref{sec:wu2021attacking}, the default implementation of \gls{apgd} (within the \autoattack library) returns adversarial points only when they are successful (i.e., when misclassification occurs).
When we use this default implementation, the \gls{hd} model which uses the WideResNet-70-16 from \citet{gowal_uncovering_2020} obtains 69.2\% $\pm$ 0.10 (compared to 65.9\% when using the points maximizing the target loss, see Table~\ref{tab:hedge-def}), which is in line with what reported by \citet{wu2021attacking} for the same classifier.
This gives us confidence that our implementation matches the one from \citeauthor{wu2021attacking}.

\subsection{Combating Adversaries with Anti-Adversaries \citep{alfarra2021combating}}

We use the official implementation of the defense, which is publicly available.\footnote{\url{https://github.com/MotasemAlfarra/Combating-Adversaries-with-Anti-Adversaries}}
For the underlying classifier, we use a WideResNet-28-10 trained with \gls{awp} \cite{wu_awp_2020}, with pre-trained weights obtained from the official AWP code repository.\footnote{\url{https://github.com/csdongxian/AWP}}
Specifically, we use the RST-AWP checkpoint, which is also referenced by \citet{alfarra2021combating} in their evaluation code.
Further following the implementation by \citet{alfarra2021combating}, we set the number of defense iterations to $N=2$ and the step size to $\alpha=0.15$.

\subsection{Online Adversarial Purification based on Self-Supervision \citep{shi2020online}}

We use the official implementation of the defense and the pre-trained model both publicly available.\footnote{\url{https://github.com/Mishne-Lab/SOAP}}
Moreover, when transferring \gls{fgsm} attacks from the static model, the defended adaptive classifier obtains a robust accuracy of 54.5\%, which is in line with what is reported by \citet{shi2020online}.

\subsection{Improving Model Robustness with Latent Distribution Locally and Globally \citep{qian2021improving}}

We use the official implementation of the defense and the pre-trained models publicly available.\footnote{\url{https://github.com/LitterQ/ATLD-pytorch}}.
All hyperparameters are set as specified by \citet{qian2021improving}.
Our only modification consists in allowing gradients to flow through the optimization steps.
We use the standard implementation of \autoattack.

\subsection{\aidpurifier: A Light Auxiliary Network for Boosting Adversarial Defense \citep{hwang2021aidpurifier}}

In our evaluation, we use the pre-trained classifiers, with WideResNet-34-10 as architecture, and discriminators, as well as the publicly available\footnote{\url{https://openreview.net/forum?id=3Uk9_JRVwiF}} original implementation of the defense.
For the parameters of \aidpurifier we use the values provided by \citet{hwang2021aidpurifier}.
Since on \svhn the defense seems to be beneficial, we further test it with \gls{apgd} with 1000 iterations and 10 restarts (divided between cross-entropy and targeted \gls{dlr} loss).
This reduces the robust accuracy to 25.0\%, suggesting that the robustness can be reduced by increasing the budget available to the attacker.

\subsection{Adversarial Attacks are Reversible with Natural Supervision \citep{mao2021adversarial}}

We use the official implementation and pre-trained classifier publicly available.\footnote{\url{https://github.com/cvlab-columbia/SelfSupDefense}}
We focus our evaluation on the \texttt{Semi-SL} model (from \citealp{carmon_unlabeled_2019}).
The original implementation of the defense retrieves negative pairs of images (to construct the contrastive loss) from the batch of images produced by the attacker.
As such, we follow the same protocol and allow the attacker to modify all the images of each batch.
Unfortunately, as the setup from \citet{mao2021adversarial} uses a batch size of 512 and requires several GPUs to work in parallel, we are forced to change the default batch size to 50, which may negatively impact the defense.
Other than that we keep all hyperparameters identical.

To evaluate whether the attacker obtains an unfair advantage by allowing it to modify the whole batch, we also re-implement our own version of the input purification procedure described by \citet{mao2021adversarial} with the negative images kept separate and hidden from the attacker.
In that setup, we were unable to improve upon the robust accuracy of the underlying static model when using transfer attacks (using \gls{apgd} on the static model) to the contrary of the full-batch setup.

\subsection{Adversarial Purification with Score-based Generative Models \citep{yoon2021adversarial}}

As base classifier we use the WideResNet-28-10 model (as evaluated by \citealp{yoon2021adversarial}) from \robustbench.
The score network is provided by \citet{song2020improved} online\footnote{\url{https://github.com/ermongroup/ncsnv2}} (we use the checkpoint named \texttt{best\_checkpoint\_with\_denoising}).
The \gls{adp} classifier attains a clean accuracy of $86.9\% \pm 0.44$ on the 1000 points used for testing robustness.
In our evaluation, we exclude the computation of the adaptive step size when getting the gradient of the target loss via backpropagation for simplicity (including it did not yield any improvements).

\section{Index of Adaptive Test-time Defenses}\label{sec:more_defenses}

Although our case study is necessarily finite, we provide further perspective by summarizing additional adaptive test-time defenses that are not included in our evaluation.
Table~\ref{tab:summary_more} indexes additional defenses known to us at publication time with the same categorization as Table~\ref{tab:summary_casestudy}.

In general, these defenses were chronologically excluded from our evaluation:
the older defenses have already been analyzed while the newer defenses have appeared concurrently or following our own work.
While each defense adapts in its own way, our case study spans their principles and building blocks, and so we hope it can guide the further examination of future adaptive test-time defenses.

\makeatletter
\setlength{\@fptop}{0pt}
\makeatother
\begin{table}[t]
\centering
\caption{%
Summary of additional adaptive test-time defenses.
We categorize each defense by its principles---input purification (IP) and model adaptation (MA)---and its building blocks: iterative algorithm (IA), auxiliary network (AN), randomization (R), and external data (ED).
Our case study covers defenses of each principle and building block but we encourage examining the full set for more perspective.
}
\label{tab:summary_more}
\vspace{2mm}
\resizebox{.5\textwidth}{!}{
\begin{tabular}{ll cc cccc}
  \multirow{2}{*}{Defense} & \multirow{2}{*}{Venue} & \multicolumn{2}{c}{Principles} & \multicolumn{4}{c}{Building blocks} \\
\cmidrule(r){3-4} \cmidrule(r){5-8}
& & IP & MA & IA & AN & R & ED \\
\toprule
\citet{song2018pixeldefend}         & ICLR    & \cmark &        & \cmark & \cmark &        &        \\
\citet{samangouei2018defensegan}    & ICLR    & \cmark &        & \cmark & \cmark & \cmark &        \\
\citet{guo_countering_2018}         & ICLR    & \cmark &        & \cmark &        & \cmark &        \\
\citet{schott2019towards}           & ICLR    &        & \cmark & \cmark &        & \cmark &        \\
\citet{wu2020rmc}                   & ICML    & \cmark &        & \cmark &        & \cmark & \cmark \\
\citet{hill2021stochastic}          & ICLR    & \cmark &        & \cmark & \cmark & \cmark &        \\
\citet{wang2021fighting}            & arXiv   &        & \cmark & \cmark &        & \cmark &        \\
\citet{nandy2021adversarially}      & arXiv   & \cmark &        & \cmark &        & \cmark &        \\
\citet{gurumurthy2021joint}         & NeurIPS &        & \cmark & \cmark &        &        &        \\
\citet{alet2021tailoring}           & NeurIPS &        & \cmark & \cmark &        & \cmark &        \\
\citet{chen2022towards}             & ICLR    &        & \cmark & \cmark &        & \cmark & \cmark \\
\citet{rusu2022hindering}           & ICML    & \cmark &        & \cmark & \cmark & \cmark &        \\
\citet{nie2022diffusion}            & ICML    & \cmark &        & \cmark & \cmark & \cmark &        \\
\citet{perez2022ags}                & ISIDA   & \cmark &        & \cmark &        &        &        \\
\bottomrule
\end{tabular}
}
\end{table}

\end{document}